\documentclass[preprint,12pt]{elsarticle}

\usepackage{physics}
\usepackage{amssymb}
\usepackage{lineno}

\usepackage{array}
\newcolumntype{L}{>{\centering\arraybackslash}m{0.05\linewidth}}

\newcolumntype{R}{>{\centering\arraybackslash}m{0.12\linewidth}}

\usepackage{textcomp,mathcomp}

\usepackage{longtable}
\usepackage{tabularray}
\UseTblrLibrary{booktabs}  

\usepackage{caption}
\usepackage{subcaption}

\usepackage[colorlinks=true,linkcolor=blue,urlcolor=black,bookmarksopen=true]{hyperref}
\usepackage{bookmark}

\usepackage{rotating}

\usepackage{adjustbox}
\usepackage{float}
\usepackage{amsmath}
\usepackage{tikz}
\usetikzlibrary{shapes,arrows}
\usepackage{algpseudocode}
\usepackage{algorithm,tabularx}
\usepackage{xcolor} 
\usepackage[T1]{fontenc}
\usepackage{pgfplots}
\usepackage{multirow}
\usepackage{siunitx}
\usepackage{threeparttable}     
\usepackage{makecell}
\usepackage{pgf-pie}

\usetikzlibrary{shapes.geometric,matrix,chains,positioning,decorations.pathreplacing,arrows,circuits.ee.IEC}

\def\HiLi{\leavevmode\rlap{\hbox to \hsize{\color{yellow!50}\leaders\hrule height .8\baselineskip depth .5ex\hfill}}}

\pgfplotsset{
	compat=newest,
	xlabel near ticks,
	ylabel near ticks
}

\tikzstyle{startstop} = [rectangle, rounded corners, minimum width=3cm, minimum height=1cm,text centered, draw=black, fill=red!30]
\tikzstyle{io} = [trapezium, trapezium left angle=70, trapezium right angle=110, minimum width=3cm, minimum height=1cm, text centered, text width=3cm, draw=black, fill=blue!30]
\tikzstyle{process} = [rectangle, minimum width=3cm, minimum height=1cm, text centered, draw=black, fill=orange!30]
\tikzstyle{decision} = [diamond, minimum width=3cm, minimum height=1cm, text centered, draw=black, fill=green!30]
\tikzstyle{arrow} = [thick,->,>=stealth]

\algnewcommand\algorithmicforeach{\textbf{for each}}
\algdef{S}[FOR]{ForEach}[1]{\algorithmicforeach\ #1\ \algorithmicdo}

\makeatletter
\newcommand{\multiline}[1]{%
  \begin{tabularx}{\dimexpr\linewidth-\ALG@thistlm}[t]{@{}X@{}}
    #1
  \end{tabularx}
}
\makeatother

\journal{TBD}

\begin{document}

\def\btc{\begin{tabular}{c}}
\def\etc{\end{tabular}}

\begin{frontmatter}

\title{A Mapping Study of Machine Learning Methods for Remaining Useful Life Estimation of Lead-Acid Batteries}

\author[add1]{S\'ergio F. Chevtchenko\corref{cor1}}
\ead{sergio.chevtchenko@sistemafiepe.org.br}
\cortext[cor1]{Corresponding author}

\author[add1]{Elisson da Silva Rocha}
\ead{elisson.rocha@sistemafiepe.org.br}

\author[add1]{Bruna Cruz}
\ead{bruna.cruz@sistemafiepe.org.br}

\author[add4]{Ermeson Carneiro de Andrade}
\ead{ermeson.andrade@ufrpe.br}

\author[add4]{Danilo Ricardo Barbosa de Araújo}
\ead{danilo.araujo@ufrpe.br}

\address[add1]{SENAI Institute of Innovation for Information and Communication Technologies (ISI-TICs), Recife, Brazil}

\address[add4]{Department of Computing at the Rural Federal University of Pernambuco (UFRPE), Recife, Brazil}

\begin{abstract}
Energy storage solutions play an increasingly important role in modern infrastructure and lead-acid batteries are among the most commonly used in the rechargeable category. Due to normal degradation over time, correctly determining the battery's State of Health (SoH) and Remaining Useful Life (RUL) contributes to enhancing predictive maintenance, reliability, and longevity of battery systems. Besides improving the cost savings, correct estimation of the SoH can lead to reduced pollution though reuse of retired batteries. This paper presents a mapping study of the state-of-the-art in machine learning methods for estimating the SoH and RUL of lead-acid batteries. 
These two indicators are critical in the battery management systems of electric vehicles, renewable energy systems, and other applications that rely heavily on this battery technology. In this study, we analyzed the types of machine learning algorithms employed for estimating SoH and RUL, and evaluated their performance in terms of accuracy and inference time.
Additionally, this mapping identifies and analyzes the most commonly used combinations of sensors in specific applications, such as vehicular batteries.  
The mapping concludes by highlighting potential gaps and opportunities for future research, which lays the foundation for further advancements in the field.

\end{abstract}


\end{frontmatter}


\section{Introduction}\label{Introduction}

The increasing demand for reliable and efficient energy storage systems has prompted significant advancements in battery technologies. Among them, lead-acid batteries have been widely used for decades due to their affordability, reliability, and high current outputs~\cite{sun2022prediction}. They play a key role in many applications, including electric vehicles, renewable energy systems, telecommunications, backup systems, and energy storage in remote areas~\cite{liu2022overview, babatunde2022assessing, vangapally2023lead}. However, like all energy storage systems, lead-acid batteries degrade over time, impacting their ability to store and deliver energy~\cite{shamim2022valve}. Thus, accurately estimating the SoH and RUL of batteries is vital for preventing unexpected failures and ensuring optimal performance.

The State of Charge (SoC) and SoH are interrelated indicators of battery performance. Both SoC and SoH play crucial roles in assessing and monitoring the performance and longevity of a battery. The SoC refers to the amount of electrical energy stored in the battery at a given time, expressed as percentage relative of the battery's full capacity. A fully charged battery is indicated as 100\%, while a fully discharged battery is denoted as 0\%. 
On the other hand, SoH estimates the battery's overall condition compared to when it was new, considering factors like charging cycles, age, and operating conditions. Similarly to SoC, SoH can be expressed as percentage relative to the battery's rated capacity~\cite{jiang2022review}. 

RUL is a critical predictive maintenance metric of a lead-acid battery. It is an estimate of the time a battery can continue operating while meeting performance requirements, considering factors like SoH, environmental conditions, and aging mechanisms. Accurately predicting RUL is challenging due to nonlinear battery degradation and the influence of factors like temperature, discharge rate, and depth of discharge~\cite{lu2017modeling}. 
The premature failure of lead-acid batteries can have serious consequences, such as power loss, disruption of essential services, and substantial costs for repairs and replacements~\cite{yahmadi2016failures}. Moreover, replacing batteries prematurely, despite being in good health, leads to resource wastage and negative environmental impacts~\cite{sun2017spent}. Hence, the development of accurate methods for estimating the SoH and RUL is crucial in order to optimize the reliability, lifespan, and efficiency of these batteries.

Advanced methods, including machine learning (ML) algorithms, are employed to integrate various factors and significantly enhance the accuracy of SoH and RUL estimates for lead-acid batteries \cite{selvabharathi2022estimating}. Therefore, the objective of this paper is to provide a comprehensive mapping study that offers an overview of recent advancements in ML methods for estimating the SoH and RUL of lead-acid batteries.
The mapping study was conducted between March and June 2023, encompassing 17 selected studies published from 2013 to 2023. Our focus was to present the ML algorithms employed, along with the associated error rates and inference times of the models. Additionally, we examined the data acquisition methods and sensors used, as well as the approaches employed to simulate or achieve battery degradation.
By analyzing the main methodologies, strengths, and limitations of these techniques, our aim is to identify gaps and opportunities for future research in this field. The findings of this study will contribute to advancing the understanding of ML-based approaches for estimating SoH and RUL in lead-acid batteries and guide future research directions.

The remainder of the paper is organized as follows.
Section~\ref{sec_related_work} presents existing reviews in the field of lead-acid batteries.
Section~\ref{ssec_study_process} introduces the protocol used to conduct the research.
Section~\ref{sec_results} presents the findings for the established research questions.
Section~\ref{sec_OpenChallenges} presents the open challenges. 
Limitations of this review and corresponding mitigation strategies are presented in Section~\ref{sec_limitations}.
Finally, Section~\ref{sec_final} presents  the conclusions, the limitations, and the future work for this study.






\section{Related Work}
\label{sec_related_work}


While there is an abundance of literature that provides mapping studies of methods for estimating battery life~\cite{lipu2018review,hasib2021comprehensive}, there is a limited emphasis on SoH and/or RUL estimation methods specifically for lead-acid batteries.

\citet{jiang2022review} conducted a review on the state of health estimation methods of lead-acid batteries.The review classified the estimation methods into four categories: direct measurement-based, model-based, data-driven, and other methods. Promising results were found with a combination of Kalman filter (KF) and data-driven methods for SoH estimation. However, accurately estimating SoH during irregular charging and discharging scenarios remained a challenge.
It is worth noting that data-driven methods include, but are not limited to, ML-based techniques. Thus, the authors  provided insights into a wide range of methodologies applied in the field, but there was no specific focus on the application of machine learning techniques.

The work by~\citet{semeraro2022battery} delved into the battery monitoring and prognostics optimization techniques. This review provided a broad understanding of model-based, data-driven and hybrid methods employed in battery health monitoring and prognostics. The analysis encompassed lead-acid, Ni-MH and lithium-ion batteries, with a significant focus on the latter.  The authors proposed  three dimensions of analysis:  battery performance, approaches, and criteria to fulfil. Similar to~\citet{jiang2022review}, data-driven methods included non-ML approaches, such as rule-based and coulumb-counting methods. The review concluded by identifying the most common model-based and data-driven approaches. A combination of both methods was suggested to compensate for the need for expert knowledge in model-based solutions and for the sensitivity to the quality and quantity of data in data-driven approaches.

\citet{pradhan2022battery} presented an important review on  battery management strategies, with a special emphasis on battery state of health monitoring techniques. While this review provided  a comprehensive study of various battery management strategies, its focus on machine learning methods for lead-acid batteries was limited, with the majority of the reviewed works focusing on lithium-ion batteries.  Additionally, machine-learning methods were considered as a subset of data-driven approaches. 

Each of these reviews provides valuable information to the field of battery health and lifespan estimation. However, the present review stands out for specifically focusing on the use of machine learning methods in estimating the state of health and remaining useful life of lead-acid batteries. It is also worth noting that the present study analyses thirteen new works that are not present in the related reviews above~\cite{olarte2022online, chen2021rapid, voronov2020predictive, voronov2018lead, frisk2015treatment, dos2016electric, luciani2022artificial, teng2023accurate, banthia2023determination, tang2019optimisation, sun2022prediction, lai2021available, selvabharathi2022estimating}. These additional works contribute to a more comprehensive and up-to-date understanding of the recent advancements and emerging trends in machine learning-based estimation methods for lead-acid batteries.

\section{Systematic Mapping Study Process}
\label{ssec_study_process}



\begin{figure}[!h]
    \centering
    \includegraphics[width=0.9\columnwidth]{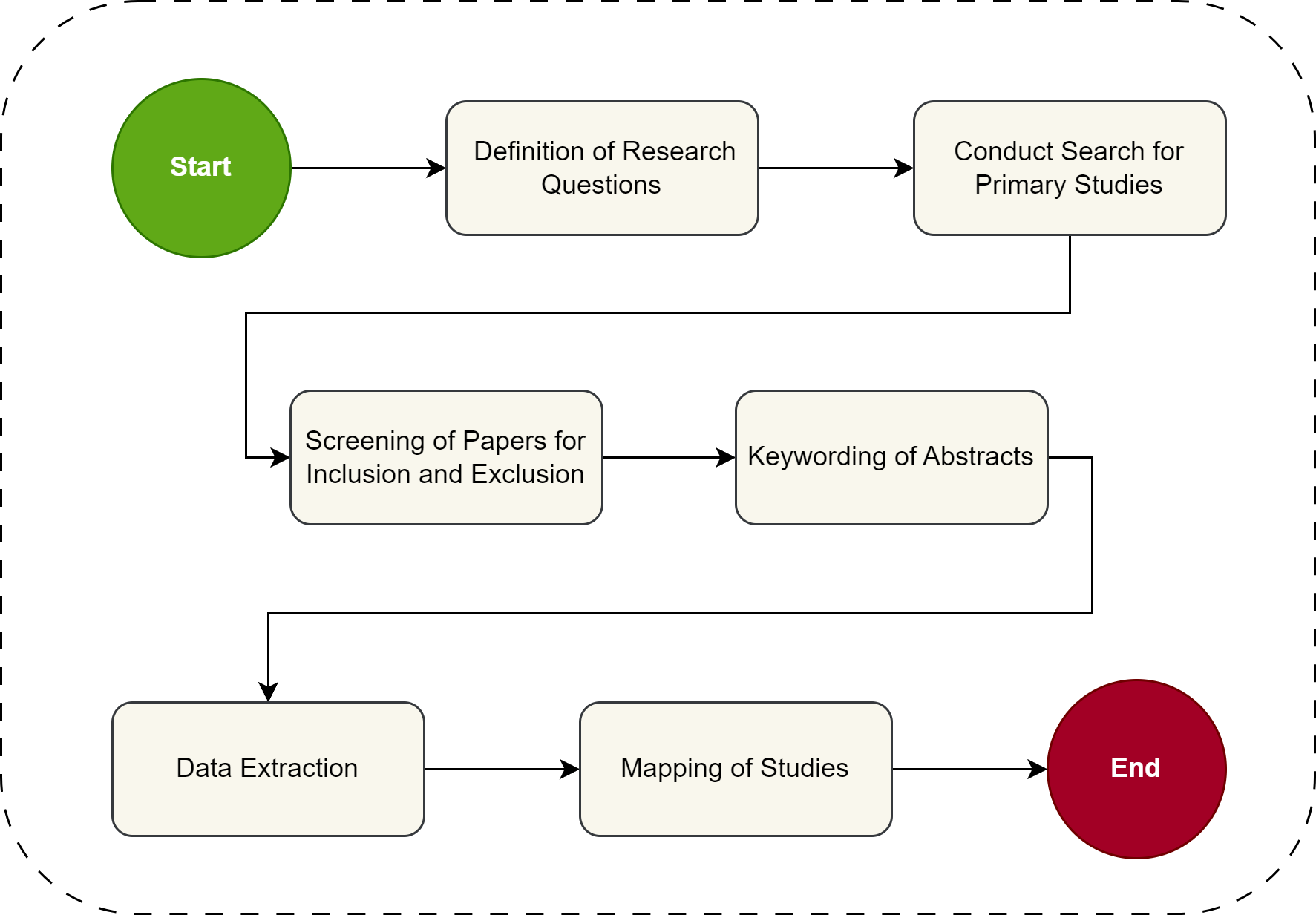}
    \caption{Steps of the adopted protocol.}
    \label{fig:methodology}
\end{figure}

Our mapping study (MS) followed the methodology outlined by Petersen \textit{et al.}~\cite{petersen2008systematic} to guide our research. This methodology allowed us to systematically identify relevant works concerning the application of machine learning techniques in estimating the remaining useful life of lead-acid batteries. 
Figure~\ref{fig:methodology} details the adapted methodology, with the steps described as follows.

\textbf{Definition of Research Questions.} Our research began by identifying important questions to investigate the current state of estimating the remaining lifespan of lead-acid batteries. We formulated the following research questions (RQs) for our study:

\begin{itemize}
    \item RQ1: Which ML algorithms are used for estimation of RUL or SoH in lead-acid batteries?
    \item RQ2: What are the average error rates and inference times for SoH or RUL estimations?
    \item RQ3: Which types of sensors are commonly used?
    \item RQ4: What type of lead-acid batteries are typically monitored?
    \item RQ5: How is the battery degradation simulated or achieved?
\end{itemize}

\textbf{Conduct Search for Primary Studies.} To conduct the search for primary studies aiming to map and address the research questions, we developed a search string for automated searches across databases. The formulation of this string involved several steps: defining relevant keywords related to the topic, exploring alternative synonyms, and combining them using logical operators such as AND and OR. The final search string used was as follows: (("lead-acid") AND ("battery") AND ("predictive maintenance" OR "remaining useful life" OR "state of health") AND ("machine learning" OR "deep learning" OR "neural network")).
The databases considered for the search of the primary studies were the ACM Digital Library\footnote{https://dl.acm.org}, IEEE Xplore\footnote{IEEExplore.ieee.org/Xplore/home.jsp}, Web of Science\footnote{https://www.webofscience.com}, Science Direct\footnote{https://www.sciencedirect.com/ } and Scopus \footnote{https://www.scopus.com/ }.

\textbf{Screening of Papers for Inclusion and Exclusion.} Given the large number of articles unrelated to our research topics, we established specific inclusion and exclusion criteria. These criteria are intended to narrow down our search and ensure that the identified literature is relevant to our evaluation search.


Our inclusion criteria were as follows: we selected studies that propose or apply ML algorithms for the estimation of RUL in lead-acid batteries. Additionally, we chose to include only peer-reviewed original studies published between 2013 and 2023.



On the other hand, our exclusion criteria consisted of: excluding articles that do not utilize ML algorithms, studies not written in English, studies with unclear results or findings, duplicated studies, articles that are not original research papers, research that does not involve the use of lead-acid batteries, as well as secondary or tertiary articles.

By applying these criteria, we aimed to ensure the relevance and quality of the studies included in our research.

\textbf{Keywording of Abstracts.} After executing the search string in the selected databases, we  proceeded with the analysis of the abstracts and metadata of the articles to verify their correspondence with the established criteria. This allowed us to determine the primary studies to be included in our review.

\textbf{Data Extraction.} With the definition of the primary studies, we extracted relevant information by carefully reading the entire articles and addressing the RQs accordingly.

\textbf{Mapping of Studies.} Finally, we presented an overview of all selected primary studies and  provided responses to each proposed RQ. This allowed us to gain insights into the current state of the literature regarding the utilization of machine learning for estimating the remaining lifespan of lead-acid batteries.

\section{Results and Discussion}
\label{sec_results}

In this section, we are conducting a descriptive analysis and addressing the research questions related to the estimation RUL and SoH in lead-acid batteries. We explore the ML algorithms used in this context, examine the average error rates and speeds of SoH or RUL estimations, investigate the data acquisition methodology and commonly used sensors in the estimation process. Additionally, we discuss the type of lead-acid batteries typically monitored and how battery degradation is simulated or achieved.

\subsection{Descriptive analysis}
\label{sec_overview}

\begin{figure}[h]
    \centering
    \includegraphics[width=0.6\columnwidth]{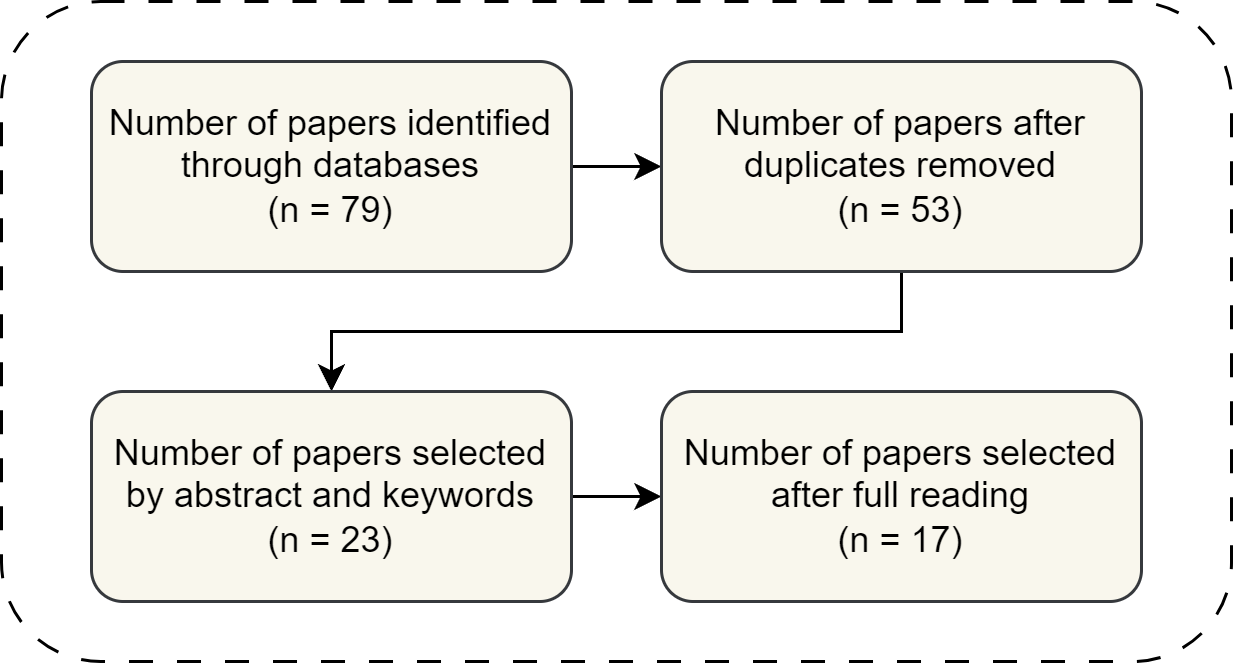}
    \caption{Flowchart of the mapping process.}
    \label{fig:flowchart}
\end{figure}

In May 2023, our search resulted in a total of 79 papers (See Figure~\ref{fig:flowchart}). 
These papers were obtained from various sources, including  7 from ACM Digital Library, 27 from Web Science, 2 from Science Direct, 34 from Scopus, and 9 from IEEE Xplore. 
After removing duplicate papers, 53 unique papers remained, which underwent a thorough screening process based on inclusion and exclusion criteria using their abstracts. 
Following this initial screening, 23 papers were selected for full-text reading and extraction of relevant information. 
Out of these, 6 papers were excluded as they did not provide direct answers to the research questions. Consequently, a total of 17 primary studies were included in the mapping process.

\begin{figure}[h]
    \centering
    \includegraphics[width=0.9\columnwidth]{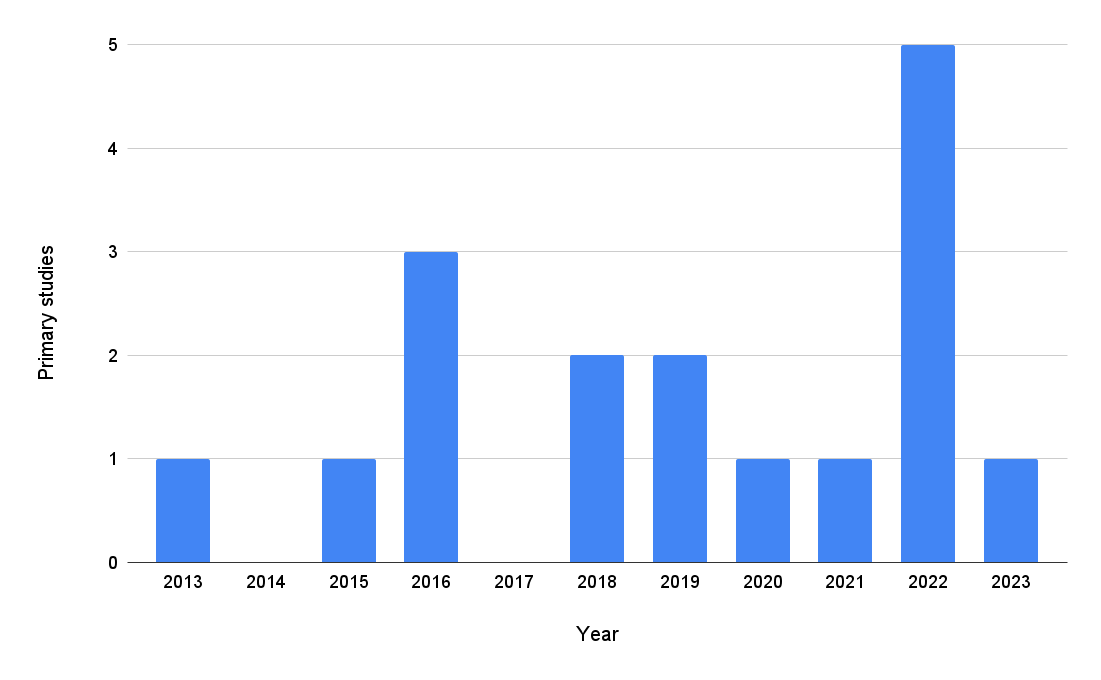}
    \caption{Number of primary studies by year.}
    \label{fig:ps_year}
\end{figure}

Out of the 17 papers selected as primary studies, 10 were obtained from the 27 articles retrieved from Web Science, 6 from the 34 articles retrieved from Scopus, and 1 from Science Direct. 
The 9 articles retrieved from IEEE Xplore were considered duplicates, as they had already been included from other sources. 
The 7 articles from the ACM Digital Library were rejected due to their focus on batteries other than lead-acid.
Figure~\ref{fig:ps_year} provides an overview of the distribution of articles in  across different years. 
The first identified article was published in 2013, which was the starting point for this research. However, there was a notable increase in the number of publications in 2016, with the inclusion of three articles. The subsequent years, 2018 and 2019, had only two articles each, but in 2022, the number increased again, with five articles meeting the inclusion criteria.




\subsection{Which ML algorithms are used for estimation of RUL or SoH in lead-acid batteries?}
\label{subsec_RQ1}
The choice of an ML algorithm, as well as its set of hyperparameters, can have a significant impact on both prediction accuracy and inference time. Therefore, in this research question, we aim to identify the most commonly used algorithms for the estimation of RUL or SoH in lead-acid batteries. An overview of ML approaches is presented in Table~\ref{tab_ml_comp_2}. 

\begin{longtblr}[
  caption = {ML Approaches, SoH Estimation Error, and Inference Time.},
  label = {tab_ml_comp_2},
]{colspec={l c c c},
rowhead = 1,
rows={font=\scriptsize}}
\toprule
\textbf{Article} & \textbf{ML algorithm} & \begin{tabular}{@{}c@{}}\textbf{Reported SoH or} \\ \textbf{RUL  error (\%)}\end{tabular} & \begin{tabular}{@{}c@{}}\textbf{Inference} \\ \textbf{time}\end{tabular}\\ 
\midrule

\citet{olarte2022online} & MLP, PSO & 0.4900 & \\ 
\citet{de2016comparison} & MLP & 1.8000 & \\ 
\citet{francca2016new} & MLP & 7.6000 & \\ 
\citet{chen2021rapid} & MLP & 1.7000 & 10 min\\ 
\citet{voronov2020predictive} & RSF, LSTM & & \\ 
\citet{voronov2018lead} & MLP & & \\ 
\citet{frisk2015treatment} & MLP & & 25s \\ 
\citet{dos2016electric} & MLP & 0.0005 & \\ 
\citet{sedighfar2018battery} & MLP & & 100 ms\\ 
\citet{talha2019neural} & MLP & & \\ 
\citet{luciani2022artificial} & MLP & & 1s\\ 
\citet{teng2023accurate} & MLP, LSTM & 2.8000 & 30 min \\ 
\citet{tang2019optimisation} & KNN & & \\ 
\citet{sun2022prediction} & BiLSTM  & 3.0000 & \\ 
\citet{lai2021available} & LSTM, MLP, RNN & 0.5800 & \\ 
\citet{shahriari2012online} & RBFNN, Fuzzy logic & 1.6000 & 2 hours\\ 
\citet{selvabharathi2022estimating} & MLP & 4.9000 & 35s\\ 

\bottomrule
\end{longtblr}

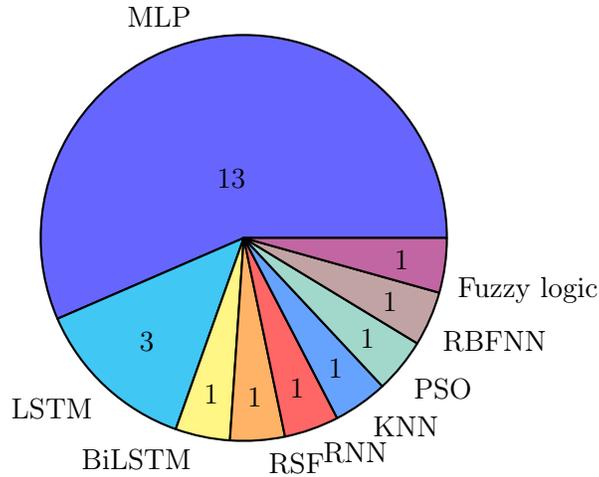
\begin{figure}[!htbp]
\centering
    \begin{tikzpicture}[scale=0.90]
    \tikzstyle{every node}=[font=\small]
    \pie[sum=auto] {
        13/MLP,
        3/LSTM,
        1/BiLSTM,
        1/RSF,
        1/RNN,
        1/KNN,
        1/PSO,
        1/RBFNN,
        1/Fuzzy logic
        }
    \end{tikzpicture}
        \caption{Number of studies using different ML algorithms.}
    \label{figure_ML_algorithms}
\end{figure}

The multilayer perceptron (MLP) is by far the most common choice, being used in 13 of the considered works (see Figure \ref{figure_ML_algorithms}). 
Interestingly, 11 of these works focused on estimating the SoH. Overall, only \citet{voronov2020predictive} and \citet{frisk2015treatment} propose a method for direct estimation of RUL. The emphasis on SoH estimation rather than RUL estimation may have been due to the nonlinear relationship between the two. Battery health can degrade at varying rates throughout its lifespan, making RUL estimation a more complex task that required additional information compared to SoH estimation. 

Rather than estimating RUL or SoH,~\citet{voronov2018lead} explored  different versions of MLPs to estimate a reliability function. This function represented the probability that a battery would last for more than a given amount of time, denoted as $t$. They found that using an ensemble of five neural networks, each with two hidden layers, led to better prediction performance. This approach outperformed single network architectures, regardless of whether they had more or fewer hidden layers.
Another study conducted by \citet{olarte2022online} compared  the performance of MLPs against the Particle Swarm Optimization (PSO) algorithm and a software-assisted human expert. They focused on identifying parameters in lead-acid batteries. The results showed that the neural network method, in this case, MLPs, achieved the highest accuracy and also demanded less computational resources compared to the other methods.

Long Short-Term Memory (LSTM) networks, which is the second most commonly used algorithm, are a type of Recurrent Neural Network (RNN), are often used for analyzing data with time dependencies. 
They are specifically designed to handle temporal sequences and capture long-range dependencies more effectively compared to traditional RNNs. LSTM stands out due to its unique structure, which includes a cell state and three types of gates: input, forget, and output. These elements enable LSTM to selectively retain or discard information, making it highly effective for tasks involving time-series data~\cite{hochreiter1997long}.

In their work,~\citet{voronov2020predictive} extended  their earlier research~\cite{voronov2018lead} by incorporating an LSTM model and comparing it with the Random Survival Forest (RSF) method. 
The aim was to estimate the potential failure time of batteries using irregular and sparse operational data obtained from vehicles, which was collected during workshop visits or through remote readouts. Due to the absence of a consistent health indicator in the data and its sparsity, they represented battery survival as a probability, referred to as a lifetime function. The authors found that an ensemble of LSTM networks yielded the best results in this particular context.

LSTM networks were also compared to MLPs for SoH estimation in~\citet{teng2023accurate}. Due to the emphasis on retired lead-acid batteries, the initial conditions were unknown, and a specific current pattern was applied for model training. Subsequent SoH estimation relied on measurements of discharge voltage and current for approximately 30 minutes. The average performance of LSTM networks was found to be slightly superior to that of MLPs, although both results were reported as being equivalent. Furthermore, an LSTM model was  proposed by~\citet{lai2021available} for estimating both the SoC and SoH specifically of gelled-electrolyte batteries used in golf carts. The study revealed  that LSTM outperformed both RNNs and MLPs when estimating SoC. Subsequently, only the LSTM model is used to evaluate SoH estimation.

Finally, the work conducted by \citet{sun2022prediction} stood out for employing a deep learning approach in the prediction of SoH. In their model, they first utilized a Convolutional Neural Network (CNN) to extract features from the data and reduce its dimensions. These processed features were then fed into a Bidirectional Long Short-Term Memory (BiLSTM) network. To enhance the learning from the time-series data, an attention mechanism was incorporated. However, it is important to note that this approach may have a drawback in terms of computational cost, as deep learning models typically require more processing power and memory compared to traditional machine learning algorithms.

\subsection{What are the average error rates and inference times for SoH or RUL estimations?}
\label{subsec_RQ5}
SoH and RUL of a battery are important factors to consider in the field of predictive maintenance. Accurately estimating the precision and inference time of these battery indicators is essential for optimizing battery performance, extending their lifespan, and ensuring their efficiency. 
In this RQ, we examine the inference time and average error rates reported in the reviewed papers. 
If a paper reported error rates for multiple ML models or configurations, we highlighted the most successful results. It is worth noting that regarding the predictions of RUL, two out of the three articles did not provide quantitative information  on precision nor inference time~\cite{voronov2020predictive, frisk2015treatment}. 

There is a substantial variation in the estimation (inference) times across different studies, ranging from seconds to hours (ver Table~\ref{tab_ml_comp_2}). For instance,~\citet{sedighfar2018battery}, reported a notably quick estimation time of roughly 100 milliseconds, although they did not numerically report the corresponding SoH estimation error.   Conversely, \citet{shahriari2012online} reported an estimation time of approximately 2 hours for charging or discharging in their study. However, the proposed method demonstrated relatively good accuracy (1.6\%) in estimating the SoH of a battery with unknown parameters.

Most studies utilized the Mean Absolute Percentage Error (MAPE) metric to report estimation errors. However, six out of the seventeen papers reviewed did not specify their estimated errors for SoH or RUL. 
Note that in some cases the error rates are reported only for SoC estimation. 
Additionally, some studies, like~\citet{luciani2022artificial}, reported errors for discrete intervals of a battery's SoH, making it challenging to include their findings in an overall comparison due to differing methodologies. 

Low reported error percentages are consistently observed in the analysed works. For instance,~\citet{lai2021available} reported a MAPE of around 0.58\%. 
It was noted that the error was closely  related to operating temperature, with higher temperatures resulting in increased estimation errors.  
Similarly,~\citet{olarte2022online} reported a low average estimation error of 0.49\%. While these studies reported the lowest error percentages in SoH estimation, it is important to note that~\citet{lai2021available} only used data on current, temperature, and voltage. 
In contrast,~\citet{olarte2022online}'s methodology was based on Electrochemical Impedance Spectroscopy (EIS). This requires the use of more expensive sensors~\cite{meng2017overview}, in addition to application of currents under specific frequencies, which can generate voltage ripples at the output of power converters~\cite{yao2021accurate}.




\subsection{Which types of sensors are commonly used?}
\label{subsec_RQ2}

The sensor types commonly used in the analyzed studies for estimating SoH or RUL of batteries are summarized in Table~\ref{tab_sens_comp}.  
The last column is reserved for sensors not associated with the three main types: voltage, current, and temperature. 
For instance, in the study by \citet{luciani2022artificial}, additional features such as capacity (Ah) and total energy (Wh) were employed. However, these features were not listed as extra sensors since they were derived from voltage and current measurements. In another case, \citet{dos2016electric} used the battery's state of charge for estimating SoH, which was calculated using the ampere-hour counting technique. This method only required a current sensor with a sufficient acquisition rate.

Table~\ref{tab_sens_comp} suggests that the most common method, utilized in 11 out of 17 reviewed studies, relies on a combination of current and voltage sensors. However, temperature fluctuations have a significant influence on SoH estimation of lead-acid batteries. As a result, five studies~\cite{chen2021rapid, sedighfar2018battery, luciani2022artificial, teng2023accurate, lai2021available} opted to use a trio of sensors: voltage, current, and temperature, aiming at a more robust SoH estimation.
Specifically for the vehicle fleet management application, a large collection of categorical and numerical data was used in~\cite{voronov2020predictive, voronov2018lead, frisk2015treatment}. In this case, the failure time of the vehicle's battery was estimated based on maintenance records and expressed as a probability function.

The study by~\citet{sun2022prediction} incorporated battery density and charging time along with voltage and temperature measurements. 
However, the specific impact of each individual feature on the results was not clearly identified, indicating the need for an additional ablation study to analyze the proposed model in more detail. Lastly, \citet{olarte2022online} adopted a model-based approach using the Electrochemical Impedance Spectroscopy (EIS) technique for battery parameter estimation. This approach involves characterizing the equivalent circuit parameters for a specific battery type and the electrochemical interpretation of these values allows for accurate inference of battery health status or failure modes. A drawback of this method is that it  involves more costly sensors and is more commonly designed for laboratory conditions~\cite{meng2017overview}.


\begin{longtblr}[
  caption = {A comparison of sensors used in selected the works.},
  label = {tab_sens_comp},
]{colspec={l c c c c},
rowhead = 1,
rows={font=\tiny}}
\toprule
\textbf{Article} & \textbf{Current} & \textbf{Voltage} & \textbf{Temperature} & \textbf{Other}\\ 
\midrule

\citet{olarte2022online} &  &  & & \begin{tabular}{@{}c@{}}Electrochemical impedance \\ spectroscopy (EIS)\end{tabular}\\ 
\citet{de2016comparison} & \checkmark &   \checkmark &     \\ 
\citet{francca2016new} & \checkmark & \checkmark & \\ 
\citet{chen2021rapid} & \checkmark & \checkmark &  \checkmark \\ 
\citet{voronov2020predictive} &  &  & & \begin{tabular}{@{}c@{}}Vehicle fleet \\ data\end{tabular}\\
\citet{voronov2018lead} &  &  & & \begin{tabular}{@{}c@{}}Vehicle fleet \\ data\end{tabular}\\
\citet{frisk2015treatment} &  & \checkmark & \checkmark & \begin{tabular}{@{}c@{}}Vehicle fleet \\ data\end{tabular}\\ 
\citet{dos2016electric} & \checkmark & \checkmark & & \\
\citet{sedighfar2018battery} & \checkmark & \checkmark & \checkmark\\
\citet{talha2019neural} & \checkmark & \checkmark & \\
\citet{luciani2022artificial} &  \checkmark & \checkmark & \checkmark & \\
\citet{teng2023accurate} & \checkmark & \checkmark & \checkmark\\
\citet{tang2019optimisation} &  \checkmark & \checkmark & \\
\citet{sun2022prediction} & & \checkmark & \checkmark & \begin{tabular}{@{}c@{}}Density \\ Charging time\end{tabular}\\ 
\citet{lai2021available} & \checkmark & \checkmark & \checkmark & Discharging time\\
\citet{shahriari2012online} & \checkmark & \checkmark & & \\
\citet{selvabharathi2022estimating} &  & \checkmark & \checkmark\\

\bottomrule
\end{longtblr}

\subsection{What type of lead-acid batteries are typically monitored?}
\label{subsec_RQ3}
\begin{figure}[!htbp]
    \centering
    \begin{tikzpicture}[scale=0.70]
    \tikzstyle{every node}=[font=\small]
    \pie[sum=auto]{
        4/VRLA,
        4/Automotive,
        3/OPzS,
        2/Retired,
        4/Generic lead-acid}
    \end{tikzpicture}
        \caption{Battery types found in the reviewed literature.}
    \label{figure_batteries_type_percentage}
\end{figure}
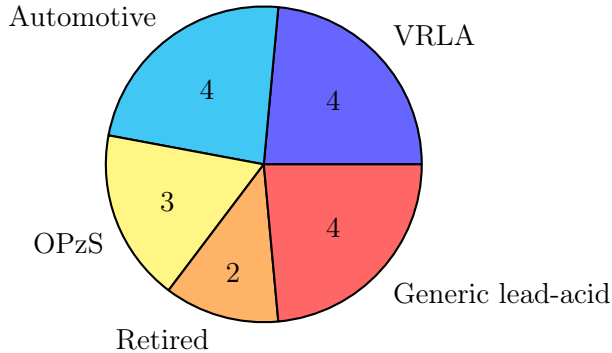

Lead-acid batteries have been widely used in various applications due to their low cost and high reliability.
With the increasing demand for efficient energy storage systems, it becomes crucial to monitor the performance and condition of different types of lead-acid batteries. 
Therefore, the objective of this research question is to  investigate the particular types of lead-acid batteries that have been predominantly monitored in the reviewed studies.
While a large portion of studies focused on generic types of lead-acid batteries without specifying the model, it is worth noting that other types, such as OPzS (tubular flooded battery)~\cite{de2016comparison, francca2016new, dos2016electric}, valve-regulated lead–acid (VRLA)~\cite{olarte2022online, lai2021available, shahriari2012online, selvabharathi2022estimating}, and automotive batteries~\cite{voronov2020predictive, voronov2018lead, frisk2015treatment, luciani2022artificial} have also been investigated. 
Figure~\ref{figure_batteries_type_percentage} illustrates the distribution of the different types of batteries identified in this mapping study.

Most approaches that studied automotive lead-acid batteries used vehicle maintenance data. For instance, in the study by~\citet{voronov2020predictive}, the batteries of approximately 2,000 vehicles were monitored for the validation/test set, and battery failures were inferred when a workshop engineer replaced them. As an exception,~\citet{luciani2022artificial} focused tests on a single automotive lead-acid battery, introducing a method for estimating battery health based on cranking current. 

Only two studies in our review used a retired type of lead-acid battery~\cite{chen2021rapid, teng2023accurate}, which brings its own set of challenges. Notably, the initial parameters of the battery under test are unknown. As a result, both the estimation of the battery's SoH and the training of the model occur offline using a dedicated testing rig. By examining the behavior and degradation patterns of 70 retired batteries, \citet{chen2021rapid} aimed to gain insight into the performance characteristics and remaining capacity of lead-acid batteries reaching the end of their operational lifespan.

The battery capacities reported in these papers range between 2.5 to 226 Ah, with an average evaluated capacity of approximately 120 Ah. However, it is worth noting that in six out of the seventeen papers, the capacity values of the investigated batteries were not explicitly stated.
Additionally, both real and simulated batteries were employed in these studies for monitoring purposes. While some works utilized real lead-acid batteries to obtain accurate and reliable data, others employed simulated batteries to model various operating conditions and evaluate the performance of monitoring systems. 
Notably,~\citet{sedighfar2018battery} was the only study that exclusively focused  on a simulated battery, with the goal of estimating the battery's SoH. This approach, however, relies heavily on knowing the battery's initial capacity to ensure accurate model training.

\subsection{How is the battery degradation simulated or achieved?}
\label{subsec_RQ4}

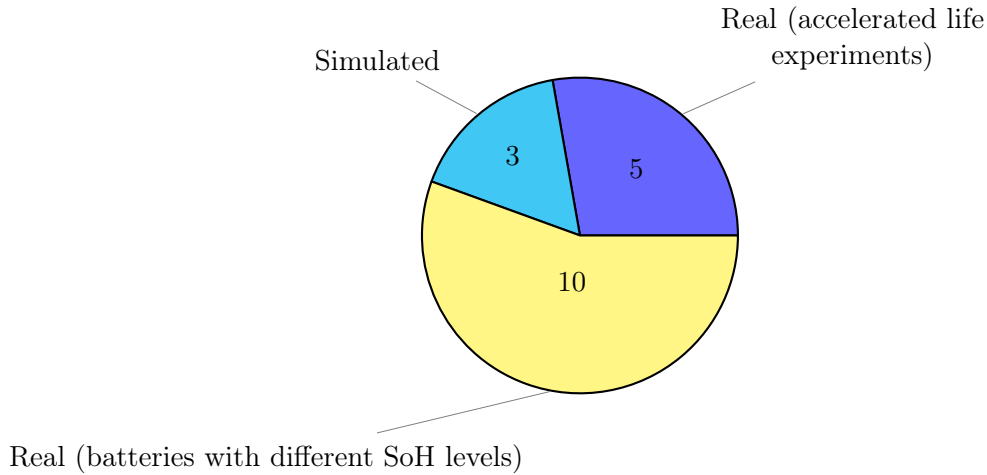
\begin{figure}[!htbp]
    \centering
    \begin{tikzpicture}[scale=0.70]
    \tikzstyle{every node}=[font=\small]
    \pie[sum=auto, /tikz/every pin/.style={align=center},
        text=pin]{
        5/Real (accelerated life  \\ experiments),
        3/Simulated,
        10/Real (batteries with different SoH levels)
      }
    \end{tikzpicture}
        \caption{Degradation methods for simulated and real batteries.}
    \label{figure_batteries_degradation_method}
\end{figure}

The simulation or achievement of battery degradation is of significant importance in the field of predictive maintenance. In this context, battery degradation is simulated or achieved using various methodologies and techniques. Experimental techniques, such as EIS and cycle tests, are  employed to characterize and quantify battery degradation. To address this question, we analyzed the degradation method reported in each primary paper. Figure~\ref{figure_batteries_degradation_method} illustrates the distribution of articles using simulated and real batteries, along with the different degradation methods employed.

The majority of the reviewed papers utilized  batteries at varying stages of health for their analysis.  
Typically, these methods tested a limited number of batteries. 
For instance,~\citet{shahriari2012online} presented different estimation techniques for analyzing battery degradation factors during charging or discharging cycles. 
However, the approach was  evaluated on just 3 VRLA batteries with varying SoH levels. 
On the other hand, the model proposed by~\citet{chen2021rapid} acquired  the complete charge and discharge data from batteries at different SoH levels and a total of 70 lead-acid batteries retired from communication base stations were used in this work.

Methods that utilized  accelerated life experiments provided a more detailed view of the algorithm's performance throughout the battery's lifespan. For example,~\citet{sun2022prediction} conducted experiments on three groups of large-capacity, flooded lead-acid batteries. Similarly,~\citet{olarte2022online} implemented a six-month aging  test with six VRLA batteries, subjecting them to various cycling and temperature operation conditions. While providing more information and robust evaluation of the proposed models, accelerate life experiments require hundreds of charge-discharge cycles, which adds to the cost of conducting such experiments.

Finally, \citet{sedighfar2018battery} relies solely on a simulated battery for experimental validation. This allows for easily changing the simulation temperature, which would present a challenge in a controlled real environment. Combination of simulated and real battery models are found in~\citet{francca2016new} and~\citet{tang2019optimisation}. In \citet{francca2016new}, simulation is used for training and preliminary testing of the MLP model, while real world data are collected for additional validation on a hybrid energy system. On the other hand, \citet{tang2019optimisation} used simulation of a hybrid energy system on a maritime vessel for estimation of energy saving when using a battery in combination with a diesel generator. 

\subsection{Discussion}
\label{subsec_discussion}

This section provides key insights from the data collected while addressing the research questions mentioned earlier. 
Most of the studies analyzed in this mapping study utilized  current, voltage, and temperature data, which indicates that these are essential variables for SoH and RUL estimation models. 
However, certain studies,  like~\citet{olarte2022online} and~\citet{sun2022prediction} achieved successful results  by incorporating additional data inputs such as EIS, battery density and charging time, respectively. It is interesting to note that some studies, including ~\citet{voronov2020predictive},~\citet{voronov2018lead}, and~\citet{frisk2015treatment}, utilized  vehicle fleet data that comprised both categorical and numerical entries. 
The inclusion of such data adds another layer of complexity to the models due to the variations in driving and charging behaviors observed across a fleet.


However, it is essential to note that not all studies have reported all the necessary information. For instance, some studies have omitted crucial details, such as inference time or the error rates associated with SoH and RUL estimation. This lack of uniform reporting presents a significant challenge when it comes to comparing results across different studies. As discussed in the Open Challenges section (see Section \ref{sec_OpenChallenges}), the field would greatly benefit from the existence of comprehensive and publicly available datasets.  

In addition to addressing the initial research questions, we also evaluated the articles in terms of three parameters, namely the specific protocol for model training, online estimation of parameters, and specific current for online estimation. The results of this evaluation can be found in Table~\ref{tab_ml_comp_1}. The specific protocol for model training indicates whether controlled experiments were conducted to generate a consistent dataset specifically for training the ML models. The online estimation of parameters evaluates the capability of the trained ML models to perform estimation of the battery's SoH or RUL without the need to disconnect the battery terminals. Lastly, the specific current for online estimation  focuses on the requirement of a specific current pattern or load condition for the estimation process to take place, even though the battery does not need to be disconnected.

\begin{longtblr}[
  caption = {Specific Protocols and Online Estimation Parameters.},
  label = {tab_ml_comp_1},
]{colspec={l c c c},
rowhead = 1,
rows={font=\tiny}}
\toprule
\textbf{Article} & \begin{tabular}{@{}c@{}}\textbf{Specific protocol for} \\ \textbf{model training}\end{tabular} & \begin{tabular}{@{}c@{}}\textbf{Online estimation} \\ \textbf{of parameters}\end{tabular} & \begin{tabular}{@{}c@{}}\textbf{Specific current for} \\ \textbf{online estimation}\end{tabular}\\ 
\midrule

\citet{olarte2022online} & \checkmark & \checkmark & \checkmark\\ 
\citet{de2016comparison} & \checkmark & \checkmark & \\ 
\citet{francca2016new} & \checkmark &  \checkmark  & \\ 
\citet{chen2021rapid} & \checkmark &  \checkmark  & \checkmark \\ 
\citet{voronov2020predictive} &  &  \checkmark & \\ 
\citet{voronov2018lead} &  &  \checkmark & \\ 
\citet{frisk2015treatment} &  & \checkmark & \\ 
\citet{dos2016electric} & \checkmark &  \checkmark  & \\ 
\citet{sedighfar2018battery} & \checkmark &  \checkmark & \\ 
\citet{talha2019neural} & \checkmark &  \checkmark & \\ 
\citet{luciani2022artificial} & \checkmark &  \checkmark & \checkmark\\ 
\citet{teng2023accurate} & \checkmark & & \\ 
\citet{tang2019optimisation} & \checkmark &  \checkmark & \checkmark\\ 
\citet{sun2022prediction} & \checkmark &  \checkmark & \checkmark\\ 
\citet{lai2021available} & \checkmark &  \checkmark & \\ 
\citet{shahriari2012online} & \checkmark &  \checkmark & \checkmark\\ 
\citet{selvabharathi2022estimating} & \checkmark & \checkmark & \checkmark\\ 

\bottomrule
\end{longtblr}

Most of the analysed studies opted  for building a dataset in a controlled environment, such as constant discharging under a fixed temperature~\cite{shahriari2012online}. 
In contrast,~\citet{voronov2020predictive},~\citet{voronov2018lead} and~\citet{frisk2015treatment} relied on data collected in the field from a large fleet of vehicles.
Online estimation was typically desirable for SoH or RUL calculation, as it eliminated the need to disconnect the battery during inference. However, this requirement was found to be less of a concern when processing retired batteries~\cite{teng2023accurate}. 
Furthermore, while most of the analyzed methods could operate online,  some necessitate required specific  current patterns~\cite{olarte2022online,luciani2022artificial}, which could be a challenging  when connected loads were  unpredictable. 



\section{Open Challenges}
\label{sec_OpenChallenges}

While a number of recent studies have utilized   machine learning for SoH and RUL estimation of lead-acid battery, there are still several open challenges that need to be addressed.  Overcoming these challenges has the potential to enhance the limitations of the existing body of work and result in more efficient algorithms. Some of the key challenges in applying machine learning to lead-acid battery SoH and RUL estimation are detailed as follows.


\textbf{Evaluation of Machine Learning Algorithms.} MLPs are commonly found in the analyzed works, likely due to their simplicity and effectiveness. 
However,  there exists a wide range of other machine learning algorithms that could be explored. Algorithms such as support vector machines (SVMs), decision trees, random forests, and more complex deep learning architectures could be better suited to certain datasets or applications. 
In particular, deep learning models such as LSTMs and transformers have achieved state-of-the-art performance in various applications involving sequential data learning.
A comprehensive comparison of these algorithms on the same dataset would be valuable in determining the best algorithm for battery SoH and RUL prediction.  It is worth noting that a publicly available dataset was not found during the present research.
 
\textbf{Automatic Hyperparameter Optimization.}
Another challenge  is the optimization of hyperparameters. Machine learning models, including MLPs, have numerous hyperparameters that need to be tuned for optimal performance. Current practices often rely on manual tuning, which is time-consuming and usually does not guarantee the optimal configuration~\cite{bergstra2011algorithms}. Automated hyperparameter optimization methods, such as Bayesian optimization and evolutionary algorithms, could significantly improve the performance of the models and should be explored in future works.

\textbf{Inference Time Evaluation.}
In real-world applications, especially in embedded systems, the speed at which a model can make a prediction could be critical. A model that takes too long to process a data stream may not be suitable for certain applications, regardless of its accuracy. 
While the evaluation of inference time was generally not conducted in the analyzed studies, future research should consider not only the accuracy of the models but also their inference time on specific hardware platforms. Assessing and optimizing inference time can be essential for ensuring the practical applicability of the models in real-time scenarios.

\textbf{Ablation Study.} 
Conducting an ablation study involves systematically removing or modifying components of the model to assess their individual impact. In the context of lead-acid battery SoH and RUL estimation, an ablation study could be highly beneficial. Specifically, investigating the influence of individual sensors on prediction accuracy could provide valuable insights. This analysis would help identify the sensors that contribute most significantly to accurate predictions, potentially leading to the development of simpler and more cost-effective battery monitoring systems.

\textbf{Hybrid Accelerated Life Experiments.}
As mentioned in Section~\ref{subsec_RQ4}, while providing more information and robust evaluation of the proposed models, accelerate life experiments require hundreds of charge-discharge cycles, which adds to the cost of conducting such experiments. However, future research could consider adopting a hybrid evaluation strategy that combines accelerated life degradation experiments with simulated models and testing on a set of pre-aged batteries. This approach would provide a more cost-effective alternative while still allowing for comprehensive evaluation of battery performance.


\textbf{Multiobjective Optimization and Feature Selection.}
There is often a trade-off between precision and computational cost when designing models.  Models with higher accuracy may require more computational resources, which can be limited in embedded applications. Multiobjective optimization techniques could be used to find a balance between precision and computational cost. Additionally, feature selection techniques could be employed to identify the most important features, reducing the dimensionality of the problem and potentially further reducing computational costs.  Incorporating these strategies can help achieve optimal performance while managing computational constraints in practical applications.

\section{Limitations of this Review}
\label{sec_limitations}

Systematic mappings are subject to risks and limitations~\cite{petersen2015guidelines}. This section presents the most common limitations and our mitigation strategies:

\par \textit{Research question formulation:} The formulation of research questions plays a crucial role in guiding the review process. However, there is a possibility of unintentionally excluding relevant studies or overlooking important aspects due to the specific wording or scope of the questions. To mitigate this, we carefully designed and refined the research questions through discussions with authors and external experts to ensure their adequacy.

\par \textit{The conduct of the search:} Despite employing a comprehensive search strategy, it is possible that some relevant studies may have been missed. This could be due to limitations in the selected databases, language restrictions, or the exclusion of certain publication types. To mitigate this limitation, we adapted our search strings for each digital database while maintaining the same terms.

\par \textit{Publication and selection bias:} The inclusion and exclusion criteria applied during the study selection process can introduce bias, as studies meeting those criteria may differ in their characteristics from those excluded. Additionally, a preference for published studies may introduce bias towards positive or significant results. 
To mitigate this limitation, we developed clear and objective inclusion and exclusion criteria. Additionally, multiple authors assessed the eligibility of the selected works to minimize subjective bias.

\par \textit{Inaccuracy in data extraction:} Misclassification or inaccuracies in data extraction refer to the potential for different reviewers to interpret the information from studies in varying ways \cite{da2014replication}. While the classification of studies is based on our judgment, there remains a possibility of incorrect categorization. To mitigate this potential issue, the classification process involved multiple author-researchers, and any discrepancies were resolved through consensus discussions.

By acknowledging these limitations and implementing appropriate strategies, we aim to provide a comprehensive and unbiased evaluation of the literature within the scope of this review.

\section{Conclusions}
\label{sec_final}

This mapping study provided an in-depth discussion on the utilization of machine learning methods for the estimation of SoH and RUL in lead-acid batteries. The review included the examination of several novel research works that have not been previously considered in similar reviews. The study was guided  by five specific research questions, which yielded insights into the commonly used ML algorithms and battery types, corresponding error rates and inference time, and identified typical experimental setups. These insights contributed to a better understanding of the current landscape of machine learning techniques for SoH and RUL estimation in lead-acid batteries.

The findings highlighted the widespread usage of  MLP and LSTM networks among the machine learning algorithms employed in this field. 
These algorithms demonstrated  effectiveness  in addressing the complexities of battery health estimation, although their performance significantly depends on sensor combination and hyperparameter tuning. The review identified  a clear need for more comprehensive evaluations of the ML algorithms, including factors such as inference time and computational cost.  Among sensors used for SoH or RUL estimation, current, voltage and temperature were identified as the most relevant. However, the impact of individual sensors is often overlooked, suggesting the need for more detailed studies in future works, including ablation and feature selection analyses. 

Furthermore, special attention was given to the online estimation capacity of each approach. 
Works that did not require the battery to be disconnected or interrupted during its normal operation were  considered suitable for a wider range of applications. Consequently, we identified  the most promising works that addressed diverse applications, such as fleet monitoring and retired battery identification.

As part of our future work, we aim to build a simulated dataset to evaluate the performance of state-of-the-art methods for SoH and RUL. This analysis will encompass the consideration of multiple conflicting parameters, including precision and computational cost. Additionally, we envision further research efforts dedicated to the development of an energy-efficient approach for inference on embedded devices.


\section*{CRediT authorship contribution statement}
Sérgio F. Chevtchenko: Writing - original draft, Conceptualization, Methodology, Investigation. Bruna Cruz: Methodology, Investigation, Writing - original draft. 
Elisson da Silva Rocha: Methodology, Writing - original draft. 
Ermeron Carneiro de Andrade: Methodology, Writing - review and editing, Supervision.
Danilo Ricardo Barbosa de Araujo: Methodology, Writing - review and editing, Supervision. 




\bibliographystyle{elsarticle-num-names}
\bibliography{ML_Batteries_review_paper}

\begin{thebibliography}{37}
\expandafter\ifx\csname natexlab\endcsname\relax\def\natexlab#1{#1}\fi
\providecommand{\url}[1]{\texttt{#1}}
\providecommand{\href}[2]{#2}
\providecommand{\path}[1]{#1}
\providecommand{\DOIprefix}{doi:}
\providecommand{\ArXivprefix}{arXiv:}
\providecommand{\URLprefix}{URL: }
\providecommand{\Pubmedprefix}{pmid:}
\providecommand{\doi}[1]{\href{http://dx.doi.org/#1}{\path{#1}}}
\providecommand{\Pubmed}[1]{\href{pmid:#1}{\path{#1}}}
\providecommand{\bibinfo}[2]{#2}
\ifx\xfnm\relax \def\xfnm[#1]{\unskip,\space#1}\fi
\bibitem[{Sun et~al.(2022)Sun, Sun, Wang, Zhou, and Cai}]{sun2022prediction}
\bibinfo{author}{S.~Sun}, \bibinfo{author}{J.~Sun}, \bibinfo{author}{Z.~Wang},
  \bibinfo{author}{Z.~Zhou}, \bibinfo{author}{W.~Cai},
\newblock \bibinfo{title}{Prediction of battery soh by cnn-bilstm network fused
  with attention mechanism},
\newblock \bibinfo{journal}{Energies} \bibinfo{volume}{15}
  (\bibinfo{year}{2022}) \bibinfo{pages}{4428}.
\bibitem[{Liu et~al.(2022)Liu, Placke, and Chau}]{liu2022overview}
\bibinfo{author}{W.~Liu}, \bibinfo{author}{T.~Placke},
  \bibinfo{author}{K.~Chau},
\newblock \bibinfo{title}{Overview of batteries and battery management for
  electric vehicles},
\newblock \bibinfo{journal}{Energy Reports} \bibinfo{volume}{8}
  (\bibinfo{year}{2022}) \bibinfo{pages}{4058--4084}.
\bibitem[{Babatunde et~al.(2022)Babatunde, Denwigwe, Oyebode, Ighravwe,
  Ohiaeri, and Babatunde}]{babatunde2022assessing}
\bibinfo{author}{O.~Babatunde}, \bibinfo{author}{I.~Denwigwe},
  \bibinfo{author}{O.~Oyebode}, \bibinfo{author}{D.~Ighravwe},
  \bibinfo{author}{A.~Ohiaeri}, \bibinfo{author}{D.~Babatunde},
\newblock \bibinfo{title}{Assessing the use of hybrid renewable energy system
  with battery storage for power generation in a university in nigeria},
\newblock \bibinfo{journal}{Environmental Science and Pollution Research}
  \bibinfo{volume}{29} (\bibinfo{year}{2022}) \bibinfo{pages}{4291--4310}.
\bibitem[{Vangapally et~al.(2023)Vangapally, Penki, Elias, Muduli, Maddukuri,
  Luski, Aurbach, and Martha}]{vangapally2023lead}
\bibinfo{author}{N.~Vangapally}, \bibinfo{author}{T.~R. Penki},
  \bibinfo{author}{Y.~Elias}, \bibinfo{author}{S.~Muduli},
  \bibinfo{author}{S.~Maddukuri}, \bibinfo{author}{S.~Luski},
  \bibinfo{author}{D.~Aurbach}, \bibinfo{author}{S.~K. Martha},
\newblock \bibinfo{title}{Lead-acid batteries and lead--carbon hybrid systems:
  A review},
\newblock \bibinfo{journal}{Journal of Power Sources} \bibinfo{volume}{579}
  (\bibinfo{year}{2023}) \bibinfo{pages}{233312}.
\bibitem[{Shamim et~al.(2022)Shamim, Viswanathan, Thomsen, Li, Reed, and
  Sprenkle}]{shamim2022valve}
\bibinfo{author}{N.~Shamim}, \bibinfo{author}{V.~V. Viswanathan},
  \bibinfo{author}{E.~C. Thomsen}, \bibinfo{author}{G.~Li},
  \bibinfo{author}{D.~M. Reed}, \bibinfo{author}{V.~L. Sprenkle},
\newblock \bibinfo{title}{Valve regulated lead acid battery evaluation under
  peak shaving and frequency regulation duty cycles},
\newblock \bibinfo{journal}{Energies} \bibinfo{volume}{15}
  (\bibinfo{year}{2022}) \bibinfo{pages}{3389}.
\bibitem[{Jiang and Song(2022)}]{jiang2022review}
\bibinfo{author}{S.~Jiang}, \bibinfo{author}{Z.~Song},
\newblock \bibinfo{title}{A review on the state of health estimation methods of
  lead-acid batteries},
\newblock \bibinfo{journal}{Journal of Power Sources} \bibinfo{volume}{517}
  (\bibinfo{year}{2022}) \bibinfo{pages}{230710}.
\bibitem[{Lu et~al.(2017)Lu, Wei, Pour, Mekonnen, and Sarwat}]{lu2017modeling}
\bibinfo{author}{J.~Lu}, \bibinfo{author}{L.~Wei}, \bibinfo{author}{M.~M.
  Pour}, \bibinfo{author}{Y.~Mekonnen}, \bibinfo{author}{A.~I. Sarwat},
\newblock \bibinfo{title}{Modeling discharge characteristics for predicting
  battery remaining life},
\newblock in: \bibinfo{booktitle}{2017 IEEE Transportation Electrification
  Conference and Expo (ITEC)}, \bibinfo{organization}{IEEE},
  \bibinfo{year}{2017}, pp. \bibinfo{pages}{468--473}.
\bibitem[{Yahmadi et~al.(2016)Yahmadi, Brik, and Ammar}]{yahmadi2016failures}
\bibinfo{author}{R.~Yahmadi}, \bibinfo{author}{K.~Brik}, \bibinfo{author}{F.~b.
  Ammar},
\newblock \bibinfo{title}{Failures analysis and improvement lifetime of lead
  acid battery in different applications},
\newblock \bibinfo{journal}{Proceedings of Engineering \& Technology (PET)}
  (\bibinfo{year}{2016}) \bibinfo{pages}{148--154}.
\bibitem[{Sun et~al.(2017)Sun, Cao, Zhang, Lin, Zheng, Cao, Sun, and
  Zhang}]{sun2017spent}
\bibinfo{author}{Z.~Sun}, \bibinfo{author}{H.~Cao}, \bibinfo{author}{X.~Zhang},
  \bibinfo{author}{X.~Lin}, \bibinfo{author}{W.~Zheng},
  \bibinfo{author}{G.~Cao}, \bibinfo{author}{Y.~Sun},
  \bibinfo{author}{Y.~Zhang},
\newblock \bibinfo{title}{Spent lead-acid battery recycling in china--a review
  and sustainable analyses on mass flow of lead},
\newblock \bibinfo{journal}{Waste Management} \bibinfo{volume}{64}
  (\bibinfo{year}{2017}) \bibinfo{pages}{190--201}.
\bibitem[{Selvabharathi and Muruganantham(2022)}]{selvabharathi2022estimating}
\bibinfo{author}{D.~Selvabharathi}, \bibinfo{author}{N.~Muruganantham},
\newblock \bibinfo{title}{Estimating the state of health of lead-acid battery
  using feed-forward neural network},
\newblock \bibinfo{journal}{Journal of Circuits, Systems and Computers}
  \bibinfo{volume}{31} (\bibinfo{year}{2022}) \bibinfo{pages}{2250081}.
\bibitem[{Lipu et~al.(2018)Lipu, Hannan, Hussain, Hoque, Ker, Saad, and
  Ayob}]{lipu2018review}
\bibinfo{author}{M.~H. Lipu}, \bibinfo{author}{M.~Hannan},
  \bibinfo{author}{A.~Hussain}, \bibinfo{author}{M.~Hoque},
  \bibinfo{author}{P.~J. Ker}, \bibinfo{author}{M.~H.~M. Saad},
  \bibinfo{author}{A.~Ayob},
\newblock \bibinfo{title}{A review of state of health and remaining useful life
  estimation methods for lithium-ion battery in electric vehicles: Challenges
  and recommendations},
\newblock \bibinfo{journal}{Journal of cleaner production}
  \bibinfo{volume}{205} (\bibinfo{year}{2018}) \bibinfo{pages}{115--133}.
\bibitem[{Hasib et~al.(2021)Hasib, Islam, Chakrabortty, Ryan, Saha, Ahamed,
  Moyeen, Das, Ali, Islam et~al.}]{hasib2021comprehensive}
\bibinfo{author}{S.~A. Hasib}, \bibinfo{author}{S.~Islam},
  \bibinfo{author}{R.~K. Chakrabortty}, \bibinfo{author}{M.~J. Ryan},
  \bibinfo{author}{D.~K. Saha}, \bibinfo{author}{M.~H. Ahamed},
  \bibinfo{author}{S.~I. Moyeen}, \bibinfo{author}{S.~K. Das},
  \bibinfo{author}{M.~F. Ali}, \bibinfo{author}{M.~R. Islam}, et~al.,
\newblock \bibinfo{title}{A comprehensive review of available battery datasets,
  rul prediction approaches, and advanced battery management},
\newblock \bibinfo{journal}{Ieee Access} \bibinfo{volume}{9}
  (\bibinfo{year}{2021}) \bibinfo{pages}{86166--86193}.
\bibitem[{Semeraro et~al.(2022)Semeraro, Caggiano, Olabi, and
  Dassisti}]{semeraro2022battery}
\bibinfo{author}{C.~Semeraro}, \bibinfo{author}{M.~Caggiano},
  \bibinfo{author}{A.-G. Olabi}, \bibinfo{author}{M.~Dassisti},
\newblock \bibinfo{title}{Battery monitoring and prognostics optimization
  techniques: challenges and opportunities},
\newblock \bibinfo{journal}{Energy}  (\bibinfo{year}{2022})
  \bibinfo{pages}{124538}.
\bibitem[{Pradhan and Chakraborty(2022)}]{pradhan2022battery}
\bibinfo{author}{S.~K. Pradhan}, \bibinfo{author}{B.~Chakraborty},
\newblock \bibinfo{title}{Battery management strategies: An essential review
  for battery state of health monitoring techniques},
\newblock \bibinfo{journal}{Journal of Energy Storage} \bibinfo{volume}{51}
  (\bibinfo{year}{2022}) \bibinfo{pages}{104427}.
\bibitem[{Olarte et~al.(2022)Olarte, Martinez~de Ilarduya, Zulueta, Ferret,
  Garcia-Ortega, and Lopez-Guede}]{olarte2022online}
\bibinfo{author}{J.~Olarte}, \bibinfo{author}{J.~Martinez~de Ilarduya},
  \bibinfo{author}{E.~Zulueta}, \bibinfo{author}{R.~Ferret},
  \bibinfo{author}{J.~Garcia-Ortega}, \bibinfo{author}{J.~M. Lopez-Guede},
\newblock \bibinfo{title}{Online identification of vlra battery model
  parameters using electrochemical impedance spectroscopy},
\newblock \bibinfo{journal}{Batteries} \bibinfo{volume}{8}
  (\bibinfo{year}{2022}) \bibinfo{pages}{238}.
\bibitem[{Chen et~al.(2021)Chen, Hsu, Lu, and Teng}]{chen2021rapid}
\bibinfo{author}{R.-J. Chen}, \bibinfo{author}{C.-W. Hsu},
  \bibinfo{author}{T.-F. Lu}, \bibinfo{author}{J.-H. Teng},
\newblock \bibinfo{title}{Rapid soh estimation for retired lead-acid
  batteries},
\newblock in: \bibinfo{booktitle}{2021 IEEE International Future Energy
  Electronics Conference (IFEEC)}, \bibinfo{organization}{IEEE},
  \bibinfo{year}{2021}, pp. \bibinfo{pages}{1--4}.
\bibitem[{Voronov et~al.(2020)Voronov, Krysander, and
  Frisk}]{voronov2020predictive}
\bibinfo{author}{S.~Voronov}, \bibinfo{author}{M.~Krysander},
  \bibinfo{author}{E.~Frisk},
\newblock \bibinfo{title}{Predictive maintenance of lead-acid batteries with
  sparse vehicle operational data},
\newblock \bibinfo{journal}{International Journal of Prognostics and Health
  Management} \bibinfo{volume}{11} (\bibinfo{year}{2020}).
\bibitem[{Voronov et~al.(2018)Voronov, Frisk, and Krysander}]{voronov2018lead}
\bibinfo{author}{S.~Voronov}, \bibinfo{author}{E.~Frisk},
  \bibinfo{author}{M.~Krysander},
\newblock \bibinfo{title}{Lead-acid battery maintenance using multilayer
  perceptron models},
\newblock in: \bibinfo{booktitle}{2018 ieee international conference on
  prognostics and health management (icphm)}, \bibinfo{organization}{IEEE},
  \bibinfo{year}{2018}, pp. \bibinfo{pages}{1--8}.
\bibitem[{Frisk and Krysander(2015)}]{frisk2015treatment}
\bibinfo{author}{E.~Frisk}, \bibinfo{author}{M.~Krysander},
\newblock \bibinfo{title}{Treatment of accumulative variables in data-driven
  prognostics of lead-acid batteries},
\newblock \bibinfo{journal}{IFAC-PapersOnLine} \bibinfo{volume}{48}
  (\bibinfo{year}{2015}) \bibinfo{pages}{105--112}.
\bibitem[{dos Santos et~al.(2016)dos Santos, de~Sousa, and
  Franca}]{dos2016electric}
\bibinfo{author}{S.~R. dos Santos}, \bibinfo{author}{T.~T. de~Sousa},
  \bibinfo{author}{A.~P. Franca},
\newblock \bibinfo{title}{Electric vehicles batteries modeling analysis based
  on a multiple layered perceptron identification approach},
\newblock in: \bibinfo{booktitle}{PCIM Europe 2016; International Exhibition
  and Conference for Power Electronics, Intelligent Motion, Renewable Energy
  and Energy Management}, \bibinfo{organization}{VDE}, \bibinfo{year}{2016},
  pp. \bibinfo{pages}{1--7}.
\bibitem[{Luciani et~al.(2022)Luciani, Feraco, Bonfitto, Amati, Tonoli, and
  Quaggiotto}]{luciani2022artificial}
\bibinfo{author}{S.~Luciani}, \bibinfo{author}{S.~Feraco},
  \bibinfo{author}{A.~Bonfitto}, \bibinfo{author}{N.~Amati},
  \bibinfo{author}{A.~Tonoli}, \bibinfo{author}{M.~Quaggiotto},
\newblock \bibinfo{title}{Artificial intelligence based state of health
  estimation with short-term current profile in lead-acid batteries for
  heavy-duty vehicles},
\newblock in: \bibinfo{booktitle}{International Design Engineering Technical
  Conferences and Computers and Information in Engineering Conference}, volume
  \bibinfo{volume}{86205}, \bibinfo{organization}{American Society of
  Mechanical Engineers}, \bibinfo{year}{2022}, p. \bibinfo{pages}{V001T01A015}.
\bibitem[{Teng et~al.(2023)Teng, Chen, Lee, and Hsu}]{teng2023accurate}
\bibinfo{author}{J.-H. Teng}, \bibinfo{author}{R.-J. Chen},
  \bibinfo{author}{P.-T. Lee}, \bibinfo{author}{C.-W. Hsu},
\newblock \bibinfo{title}{Accurate and efficient soh estimation for retired
  batteries},
\newblock \bibinfo{journal}{Energies} \bibinfo{volume}{16}
  (\bibinfo{year}{2023}) \bibinfo{pages}{1240}.
\bibitem[{Banthia et~al.(2023)Banthia, Paliwal, Patle, Ghole, and
  Paliwal}]{banthia2023determination}
\bibinfo{author}{A.~Banthia}, \bibinfo{author}{S.~Paliwal},
  \bibinfo{author}{V.~Patle}, \bibinfo{author}{M.~S. Ghole},
  \bibinfo{author}{P.~Paliwal},
\newblock \bibinfo{title}{Determination of battery life using state of health
  data and linear regression},
\newblock in: \bibinfo{booktitle}{2023 IEEE International Students' Conference
  on Electrical, Electronics and Computer Science (SCEECS)},
  \bibinfo{organization}{IEEE}, \bibinfo{year}{2023}, pp.
  \bibinfo{pages}{1--5}.
\bibitem[{Tang et~al.(2019)Tang, Dickie, Roman, Robu, and
  Flynn}]{tang2019optimisation}
\bibinfo{author}{W.~Tang}, \bibinfo{author}{R.~Dickie},
  \bibinfo{author}{D.~Roman}, \bibinfo{author}{V.~Robu},
  \bibinfo{author}{D.~Flynn},
\newblock \bibinfo{title}{Optimisation of hybrid energy systems for maritime
  vessels},
\newblock \bibinfo{journal}{The Journal of Engineering} \bibinfo{volume}{2019}
  (\bibinfo{year}{2019}) \bibinfo{pages}{4516--4521}.
\bibitem[{Lai and Kuo(2021)}]{lai2021available}
\bibinfo{author}{C.-M. Lai}, \bibinfo{author}{T.-J. Kuo},
\newblock \bibinfo{title}{Available capacity computation model based on long
  short-term memory recurrent neural network for gelled-electrolyte batteries
  in golf carts},
\newblock \bibinfo{journal}{IEEE Access} \bibinfo{volume}{10}
  (\bibinfo{year}{2021}) \bibinfo{pages}{54433--54444}.
\bibitem[{Petersen et~al.(2008)Petersen, Feldt, Mujtaba, and
  Mattsson}]{petersen2008systematic}
\bibinfo{author}{K.~Petersen}, \bibinfo{author}{R.~Feldt},
  \bibinfo{author}{S.~Mujtaba}, \bibinfo{author}{M.~Mattsson},
\newblock \bibinfo{title}{Systematic mapping studies in software engineering},
\newblock in: \bibinfo{booktitle}{12th International Conference on Evaluation
  and Assessment in Software Engineering (EASE) 12}, \bibinfo{year}{2008}, pp.
  \bibinfo{pages}{1--10}.
\bibitem[{De~Sousa et~al.(2016)De~Sousa, Arioli, Vieira, dos Santos, and
  Fran{\c{c}}a}]{de2016comparison}
\bibinfo{author}{T.~T. De~Sousa}, \bibinfo{author}{V.~T. Arioli},
  \bibinfo{author}{C.~S. Vieira}, \bibinfo{author}{S.~R. dos Santos},
  \bibinfo{author}{A.~P. Fran{\c{c}}a},
\newblock \bibinfo{title}{Comparison of different approaches for lead acid
  battery state of health estimation based on artificial neural networks
  algorithms},
\newblock in: \bibinfo{booktitle}{2016 IEEE Conference on Evolving and Adaptive
  Intelligent Systems (EAIS)}, \bibinfo{organization}{IEEE},
  \bibinfo{year}{2016}, pp. \bibinfo{pages}{79--84}.
\bibitem[{Fran{\c{c}}a et~al.(2016)Fran{\c{c}}a, de~Sousa, Arioli, dos Santos,
  Rosolem, de~Castro, do~Nascimento, and Vieira}]{francca2016new}
\bibinfo{author}{A.~Fran{\c{c}}a}, \bibinfo{author}{T.~T. de~Sousa},
  \bibinfo{author}{V.~T. Arioli}, \bibinfo{author}{S.~R. dos Santos},
  \bibinfo{author}{M.~F. Rosolem}, \bibinfo{author}{P.~C. de~Castro},
  \bibinfo{author}{T.~C. do~Nascimento}, \bibinfo{author}{C.~S. Vieira},
\newblock \bibinfo{title}{A new approach to estimate soh of lead-acid batteries
  used in off-grid pv system},
\newblock in: \bibinfo{booktitle}{2016 IEEE International Telecommunications
  Energy Conference (INTELEC)}, \bibinfo{organization}{IEEE},
  \bibinfo{year}{2016}, pp. \bibinfo{pages}{1--7}.
\bibitem[{Sedighfar and Moniri(2018)}]{sedighfar2018battery}
\bibinfo{author}{A.~Sedighfar}, \bibinfo{author}{M.~Moniri},
\newblock \bibinfo{title}{Battery state of charge and state of health
  estimation for vrla batteries using kalman filter and neural networks},
\newblock in: \bibinfo{booktitle}{2018 5th International Conference on
  Electrical and Electronic Engineering (ICEEE)}, \bibinfo{organization}{IEEE},
  \bibinfo{year}{2018}, pp. \bibinfo{pages}{41--46}.
\bibitem[{Talha et~al.(2019)Talha, Asghar, and Kim}]{talha2019neural}
\bibinfo{author}{M.~Talha}, \bibinfo{author}{F.~Asghar}, \bibinfo{author}{S.~H.
  Kim},
\newblock \bibinfo{title}{A neural network-based robust online soc and soh
  estimation for sealed lead--acid batteries in renewable systems},
\newblock \bibinfo{journal}{Arabian Journal for Science and Engineering}
  \bibinfo{volume}{44} (\bibinfo{year}{2019}) \bibinfo{pages}{1869--1881}.
\bibitem[{Shahriari and Farrokhi(2012)}]{shahriari2012online}
\bibinfo{author}{M.~Shahriari}, \bibinfo{author}{M.~Farrokhi},
\newblock \bibinfo{title}{Online state-of-health estimation of vrla batteries
  using state of charge},
\newblock \bibinfo{journal}{IEEE Transactions on Industrial Electronics}
  \bibinfo{volume}{60} (\bibinfo{year}{2012}) \bibinfo{pages}{191--202}.
\bibitem[{Hochreiter and Schmidhuber(1997)}]{hochreiter1997long}
\bibinfo{author}{S.~Hochreiter}, \bibinfo{author}{J.~Schmidhuber},
\newblock \bibinfo{title}{Long short-term memory},
\newblock \bibinfo{journal}{Neural computation} \bibinfo{volume}{9}
  (\bibinfo{year}{1997}) \bibinfo{pages}{1735--1780}.
\bibitem[{Meng et~al.(2017)Meng, Ricco, Luo, Swierczynski, Stroe, Stroe, and
  Teodorescu}]{meng2017overview}
\bibinfo{author}{J.~Meng}, \bibinfo{author}{M.~Ricco},
  \bibinfo{author}{G.~Luo}, \bibinfo{author}{M.~Swierczynski},
  \bibinfo{author}{D.-I. Stroe}, \bibinfo{author}{A.-I. Stroe},
  \bibinfo{author}{R.~Teodorescu},
\newblock \bibinfo{title}{An overview and comparison of online implementable
  soc estimation methods for lithium-ion battery},
\newblock \bibinfo{journal}{IEEE Transactions on Industry Applications}
  \bibinfo{volume}{54} (\bibinfo{year}{2017}) \bibinfo{pages}{1583--1591}.
\bibitem[{Yao et~al.(2021)Yao, Lu, and Lei}]{yao2021accurate}
\bibinfo{author}{Q.~Yao}, \bibinfo{author}{D.-D.-C. Lu},
  \bibinfo{author}{G.~Lei},
\newblock \bibinfo{title}{Accurate online battery impedance measurement method
  with low output voltage ripples on power converters},
\newblock \bibinfo{journal}{Energies} \bibinfo{volume}{14}
  (\bibinfo{year}{2021}) \bibinfo{pages}{1064}.
\bibitem[{Bergstra et~al.(2011)Bergstra, Bardenet, Bengio, and
  K{\'e}gl}]{bergstra2011algorithms}
\bibinfo{author}{J.~Bergstra}, \bibinfo{author}{R.~Bardenet},
  \bibinfo{author}{Y.~Bengio}, \bibinfo{author}{B.~K{\'e}gl},
\newblock \bibinfo{title}{Algorithms for hyper-parameter optimization},
\newblock \bibinfo{journal}{Advances in neural information processing systems}
  \bibinfo{volume}{24} (\bibinfo{year}{2011}).
\bibitem[{Petersen et~al.(2015)Petersen, Vakkalanka, and
  Kuzniarz}]{petersen2015guidelines}
\bibinfo{author}{K.~Petersen}, \bibinfo{author}{S.~Vakkalanka},
  \bibinfo{author}{L.~Kuzniarz},
\newblock \bibinfo{title}{Guidelines for conducting systematic mapping studies
  in software engineering: An update},
\newblock \bibinfo{journal}{Information and software technology}
  \bibinfo{volume}{64} (\bibinfo{year}{2015}) \bibinfo{pages}{1--18}.
\bibitem[{Da~Silva et~al.(2014)Da~Silva, Suassuna, Fran{\c{c}}a, Grubb,
  Gouveia, Monteiro, and dos Santos}]{da2014replication}
\bibinfo{author}{F.~Q. Da~Silva}, \bibinfo{author}{M.~Suassuna},
  \bibinfo{author}{A.~C.~C. Fran{\c{c}}a}, \bibinfo{author}{A.~M. Grubb},
  \bibinfo{author}{T.~B. Gouveia}, \bibinfo{author}{C.~V. Monteiro},
  \bibinfo{author}{I.~E. dos Santos},
\newblock \bibinfo{title}{Replication of empirical studies in software
  engineering research: a systematic mapping study},
\newblock \bibinfo{journal}{Empirical Software Engineering}
  \bibinfo{volume}{19} (\bibinfo{year}{2014}) \bibinfo{pages}{501--557}.

\end{thebibliography}

\end{document}